\documentclass[10pt,twocolumn,letterpaper]{article}

\usepackage{cvpr}              % To produce the CAMERA-READY version
\usepackage{amssymb}
\usepackage{siunitx}
\usepackage{booktabs}
\usepackage{dcolumn}
\newcolumntype{d}[1]{D{:}{\,:\,}{#1}}
\usepackage{caption}

\usepackage[dvipsnames]{xcolor}

\definecolor{cvprblue}{rgb}{0.21,0.49,0.74}
\usepackage[pagebackref,breaklinks,colorlinks,citecolor=cvprblue]{hyperref}

\def\paperID{13710} % *** Enter the Paper ID here
\def\confName{CVPR}
\def\confYear{2024}

\title{SA-MixNet: Structure-aware Mixup and Invariance Learning 

for Scribble-supervised Road Extraction in Remote Sensing Images}

\author{
Jie~Feng$^{1}$\thanks{Corresponding author: jiefeng@xidian.edu.cn} \quad Hao~Huang$^{1}$ \quad Junpeng~Zhang$^{1}$ \quad Weisheng~Dong$^{1}$ \quad Dingwen~Zhang$^{2}$ \quad Licheng~Jiao$^{1}$\\
\\
$^{1}$Xidian University\\
{\tt\small jiefeng@xidian.edu.cn}\\
$^{2}$Northwestern Polytechnical University\\
}

\begin{document}
\maketitle
\begin{abstract}

Mainstreamed weakly supervised road extractors rely on highly confident pseudo-labels propagated from scribbles, and their performance often degrades gradually as the image scenes tend various.
We argue that such degradation is due to the poor model's invariance to scenes with different complexities, whereas existing solutions to this problem are commonly based on crafted priors that cannot be derived from scribbles.
To eliminate the reliance on such priors, we propose a novel Structure-aware Mixup and Invariance Learning framework (SA-MixNet) for weakly supervised road extraction that improves the model invariance in a data-driven manner.
Specifically, we design a structure-aware Mixup scheme to paste road regions from one image onto another for creating an image scene with increased complexity while preserving the road's structural integrity.
Then an invariance regularization is imposed on the predictions of constructed and origin images to minimize their conflicts, which thus forces the model to behave consistently on various scenes.
Moreover, a discriminator-based regularization is designed for enhancing the connectivity meanwhile preserving the structure of roads.
Combining these designs, our framework demonstrates superior performance on the DeepGlobe, Wuhan, and Massachusetts datasets outperforming the state-of-the-art techniques by 1.47\%, 2.12\%, 4.09\% respectively in IoU metrics, and showing its potential of plug-and-play. The code will be made publicly available.

\end{abstract}    
\section{Introduction}
\label{sec:intro}

% In recent years, research in the field of road extraction has made remarkable progress.
% However, it heavily relies on extensive and fine-grained annotation data.
% The scarcity of precise labeling greatly restricts the full utilization of remote sensing images,  stimulating a multitude of studies aiming to interpret remote sensing images with fewer fully annotated data.  Specifically, semi-supervised semantic segmentation (SSS)\cite{FMWDCT} which still needs some precise annotations, and weakly supervised semantic segmentation (WSS) using scribble\cite{2019RoadScribble, 2021Scribble}, or point\cite{2021WeaklyPoint} label.

\newcounter{example1}
\setcounter{example1}{1}

\newcounter{example2}
\setcounter{example2}{2}

\newcounter{example3}
\setcounter{example3}{3}

\begin{figure}[h] 
\centering
\includegraphics[width=0.48\textwidth]{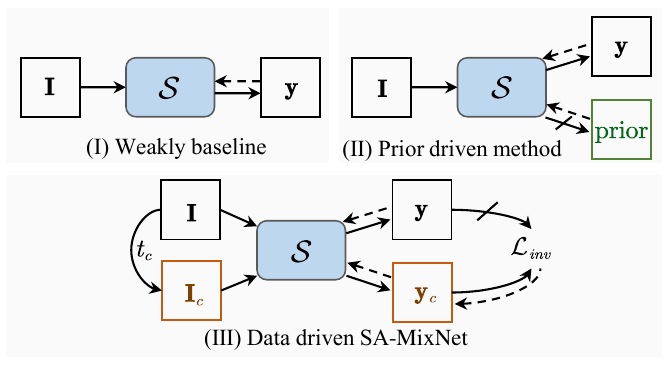} 
\caption{Overview of weakly supervised road extraction architectures:(\Roman{example1})Weakly supervised baseline with pseudo segmentation loss only, (\Roman{example2})Prior-driven method with additional manual prior, (\Roman{example3})Our proposed framework from data-driven manner without additional prior. Where $t_c$ represents the sample construction. `$\rightarrow$' means the forward operation. `$\dashrightarrow$' means backpropagation. `/' on `$\rightarrow$' means stop-gradient. $\mathcal{L}_{inv}$ means invariance regularization.}
\label{fig_intro} 
\end{figure}
% 大纲
% 全监督道路提取任务取得重大进展，然而受限于精确标注，应用场景受限
In recent years, remarkable performance improvement has been achieved in road extraction \cite{2018Stacked, 2021CoANet, 2022_split, 2020topology}, which partly relies on detailed annotations.
In practical applications, it is challenging and costly to obtain such annotations, especially on large-scale remote sensing images.
Therefore, learning with limited annotations has been emerging as a practical solution in such situations.
Despite that point annotation has been continuously explored for weakly supervised road extraction \cite{2021WeaklyPoint}, scribble is recognized as a better annotation form \cite{2019RoadScribble, 2021Scribble, Zhou_Sui_Chen_Liu_Shi_Chen} as it provides additional clues on road structure.

In order to learn from scribbles, the expansion of highly confident supervision areas is key.
% Road extraction with scribbles commonly follows a paradigm that propagates pseudo-labels from scribbles and then learns from labels with the noise.
% In road extraction, the generation of pseudo-labels is key for effective learning from scribble.
Statistic-based methods \cite{2019RoadScribble, Zhou_Sui_Chen_Liu_Shi_Chen} employ a predefined width to expand the scribbles guided by statistic knowledge.
However, a conservative width setting is necessary to avoid introducing incorrect information when dealing with significant variations in road widths, resulting in a lack of supervision. 
On a different front, methods focusing on image content \cite{2021Scribble} aim to reduce the reliance on statistics and handle the variety of road's width, cluster similar pixel regions, and propagate labels by using image features \cite{Sinop_Grady_2007}. 
%这句什么意思
% However, the structure characteristics of roads are not considered in these methods, leading to unnecessary unlabeled areas.
Due to the sensitivity to the initial clustering points, the clustering regions are difficult to adhere to rough road boundaries, which causes noisy annotations.
%%下面这个总结也不太好
In all cases, limitations and noise in annotations require additional regularized supervision to improve model generalization ability.

% Appropriate regularization design is vital to learn from the pseudo-labels with noise.
Some attempts have been made by introducing strong priors to the regularization learning process, shown in \cref{fig_intro}.
%这名字有问题，基于损失的正则这个
Graph-based regularization methods \cite{2019RoadScribble,2021Scribble,Cut_Loss,CRF_Loss} are commonly used based on pixel similarity and structural priors, but they are sensitive to noise and high complexity in remote sensing images. 
On the other hand, collaborative learning-based regularization methods \cite{2021Scribble, Zhou_Sui_Chen_Liu_Shi_Chen} focus on road surface as well as additional edge or direction prior supervision, which can not be derived from weak labels. 
Additionally, existing methods' performance commonly degrades as the samples tend to be complex in scenes of intertwined roads and indistinguishable backgrounds, leading to more missed detections in regions of confusion.

We argue that such degradation is due to the model's poor invariance to road targets in scenes with different complexities.
To address the above challenges, we proposed SA-MixNet, a novel Structure-aware Mixup and Invariance Learning framework for weakly supervised road extraction via scribble labels, shown in \cref{fig_intro}.
First, we propose an improved mixup method, SA-Mixup, to operate on a pair of training samples by pasting one's foreground regions onto the other. It constructs difficult samples with complex scenes while maintaining the road's structural integrity.
Then, to approach consistent performance on both constructed difficult and original samples, the invariance regularization is imposed on their predictions to minimize the conflict.
It makes the model learn more robust and invariant foreground features by bridging the performance gap between difficult samples and general samples.
Moreover, a topology regularization based on GAN is added to enhance the roads' connectivity.
To guarantee the performance of sample construction, a Statistic and Content-based Label Expansion (SCLE) is proposed to expand the highly confident foreground areas by taking road structure into specific consideration.

The main contributions of this paper are as follows:

\begin{enumerate}

\item

We propose SA-MixNet, a novel framework for weakly supervised road extraction via scribble labels, that enhances the model's generalization ability from a data-driven perspective and eliminates the requirements for additional priors.

% \item

% We propose a Prior and Content-based Label Expansion (PCLE) scheme that fuses prior-based and road-specific content-based label masks, balancing the contradiction between the completeness and accuracy of pseudo-label generation.

\item

The proposed structure-aware Mixup constructs proper scenes with various complexities for the invariance learning framework while mitigating road sample imbalance and maintaining the road's structural integrity.

\item

The invariance regularization is proposed to force the model to approach consistent performance on original and constructed samples. It guarantees the model's invariance under different conditions, resulting in a profound performance improvement on ambiguity samples.

\item

Experiments on three datasets \emph{i.e.} DeepGlobe, Wuhan, and Massachusetts-road, show that SA-MixNet achieves the best performance on the weakly supervised road extraction, and shows the potential of its plug-and-play to be used to improve the performance of other existing methods.

\end{enumerate}

\begin{figure*}[h] 
\centering
\includegraphics[width=0.9\textwidth]{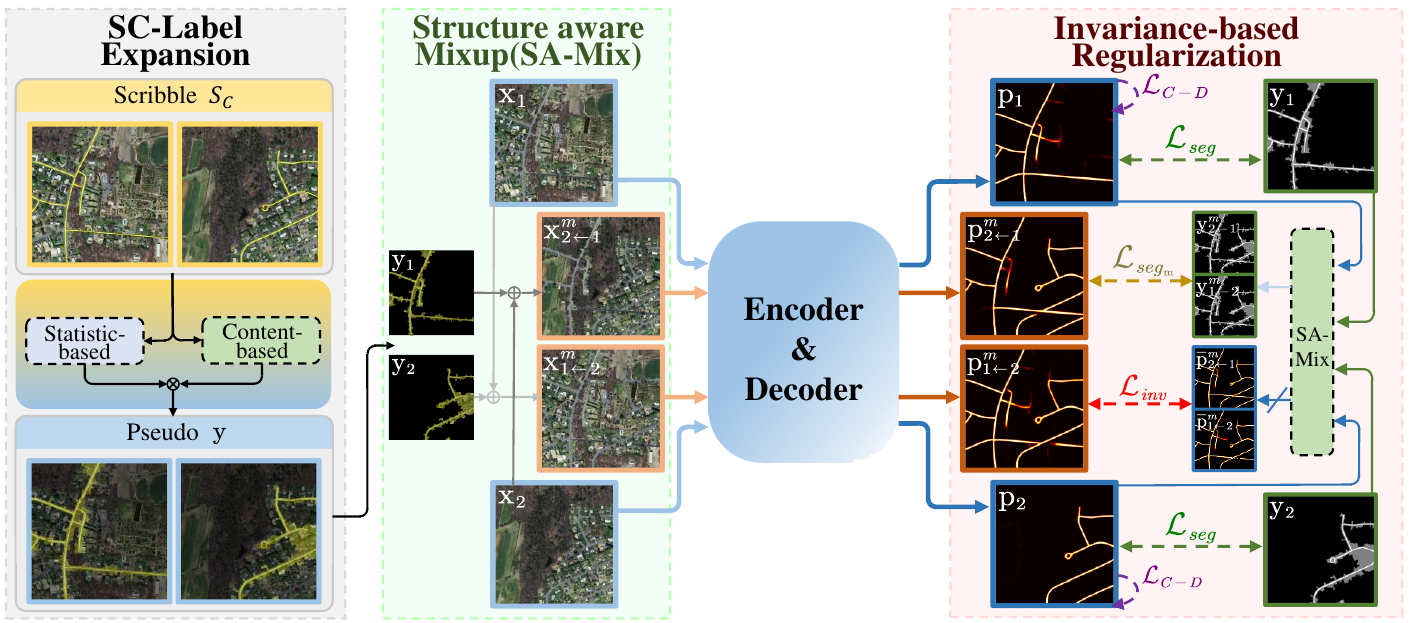} 
\caption{The pipeline of proposed SA-MixNet, consisting of three parts: Statistic and Content-based (SC) Label Expansion, Structure-aware Mixup (SA-Mix) based sample construction, and Invariance-based Regularization including base segmentation loss $(\mathcal{L}_{seg} \ \& \ \mathcal{L}_{seg_\text{m}})$, invariance regularization $(\mathcal{L}_{inv})$, and the connectivity regularization $(\mathcal{L}_{C-D})$. `/' on `$\rightarrow$' means stop-gradient.}
\label{fig_overview} 
\end{figure*}
\section{Related Works}
\label{SC:Related Works}

\noindent
\subsection{Weakly Supervised Road Extraction}

% 自然图像语义分割分为迭代法和非迭代法，各有何优缺点？
% In the field of weakly supervised semantic segmentation, there are primarily two approaches: iterative methdos\cite{Wei_Feng_Liang_Cheng_Zhao_Yan_2017, Durand_Mordan_Thome_Cord_2017, Ahn_Kwak_2018}, and none-iterative methods\cite{Random_Walk, Bai_Sapiro_2007, Sinop_Grady_2007}. 
% Iterative methods generate pseudo labels by previous round model's predictions, which often suffer from the cumulative effect of errors. 
% While the non-iterative methods rely on the preprocessing of weak labels and expanding labeled regions through propagation, which is difficult for the complex edges of natural images.

%为什么道路常采取非迭代法
% However, road targets in remote sensing images typically exhibit the characteristic of being equally wide on both sides and parallel, allowing for the design of simple methods to achieve efficient label propagation. Therefore, non-iterative methods are the mainstream in road extraction. 

%道路领域生成伪标签主要方式
% In the field of weakly supervised semantic segmentation, there are primarily two approaches: iterative\cite{Wei_Feng_Liang_Cheng_Zhao_Yan_2017, Durand_Mordan_Thome_Cord_2017, Ahn_Kwak_2018}, and none-iterative methods\cite{Random_Walk, Bai_Sapiro_2007, Sinop_Grady_2007}. 
% In the context of weak labels, since scribbles possess structural features most similar to roads and their characteristics parallel to road edges, algorithms for generating pseudo-labels based on scribbles can achieve relatively good performance. Therefore, most existing algorithms are non-iterative methods based on scribbles.

In road extraction from remote sensing images, imbalanced sample distribution and undistinguishable roads from background often lead to broken connections and incomplete structures, especially in complex scenes.
Such difficulties tend more severe under limited annotations. 
Existing weakly supervised methods attempt to mitigate this issue by focusing on pseudo-label generation and regularization learning.

In the label generation phase, statistics-based methods \cite{2019RoadScribble, Zhou_Sui_Chen_Liu_Shi_Chen} can simply and effectively extend scribbles but may struggle to accommodate roads of varying widths. 
Wei et al.\cite{2021Scribble} used SLIC \cite{SLIC} clustering to combine statistical and image features, reducing reliance on statistical knowledge. 
However, SLIC's sensitivity to initial points leads to noisy annotations to road boundaries.

Learning from such limited annotations with noise requires additional regularization supervision to enhance the model's generalization performance. 
Some methods provide extra supervision based on graph optimization \cite{2019RoadScribble, Cut_Loss, 2021Scribble, CRF_Loss} through pixel similarity and structural priors, but they are susceptible to inherent noise in remote sensing images and perform poorly in complex scenes, affecting the model's generalization.
Other methods use multi-branch networks \cite{2021Scribble, Zhou_Sui_Chen_Liu_Shi_Chen}  to extract road surface in a collaborative learning manner, but they typically require additional edge or direction prior supervision that cannot be derived from scribbles.

\subsection{Mixup Strategies}
The addition of perturbation on input-level \cite{2016Regularization, 2022Learning, 2022CycleMix, Tack_Yu_Jeong_Kim_Hwang_Shin_2022}, feature-level \cite{2020Semi, Rasmus_Valpola_Honkala_Berglund_Raiko_2015, Xu_Zeng_Lian_Ding_2022} or both \cite{2016Temporal, 2022Learning} has been used in semi-supervised and self-supervised learning to generalize models with limited data.    
Input-level perturbation improves the robustness and generalizability of both encoders and decoders by utilizing unlabeled data efficiently. Its simplicity and transferability have contributed to its broad application. Recently, mixup techniques stand out as an effective method for implementing input-level perturbation.

Rooted in data augmentation basics like rotation, cropping, flipping, and color jittering,  non-heuristic Mixup  \cite{2017_mixup} blends the images and labels by pixel-wise linear addition to craft new examples, improving model performance with subtle data variations. 
Other non-heuristic methods like CutMix\cite{2019_cutmix} and FMix\cite{FMix} randomly generate rectangular or smooth masks for pasting and covering between samples. 
Building on this, heuristic methods such as Co-Mixup\cite{Co-Mixup} and PuzzleMix \cite{2020_puzzlemix} preserve areas of high responsiveness during pasting based on the strength of gradients from backpropagation. 

For road targets, most of the existing mixup methods select partial image blocks for pasting without considering the grid-like characteristics and spanning nature of road objects in remote sensing images, bringing damage to roads' structure integrity.
\section{Methodology}
\label{sc:method}
\subsection{Overview of SA-MixNet}
The proposed SA-MixNet presents a unified weakly supervised framework for road extraction using scribble labels.
As illustrated in \cref{fig_overview}, it includes the basic statistic and content-based label expansion, structure-aware Mixup module, and invariance regularization. 
The label propagation first updates the scribble labels with statistic information, and refined statistic-based pseudo label by incorporating road-specific content-based clustering into statistic-based expansion. 
Then, structure-aware Mixup module is proposed by pasting the generated roads and buffers randomly to construct new image scenes with various complexities. 
Finally, the invariance regularization is imposed to force the model to approach consistent performance on original and constructed samples.
Moreover, a discriminator-based regularization is added for enhancing the connectivity meanwhile preserving the structure of the road.

\subsection{Statistic and Content-based Label Expansion}
\label{sc:RoadLabel}
\begin{figure}[h] 
\centering
\includegraphics[width=0.45\textwidth]{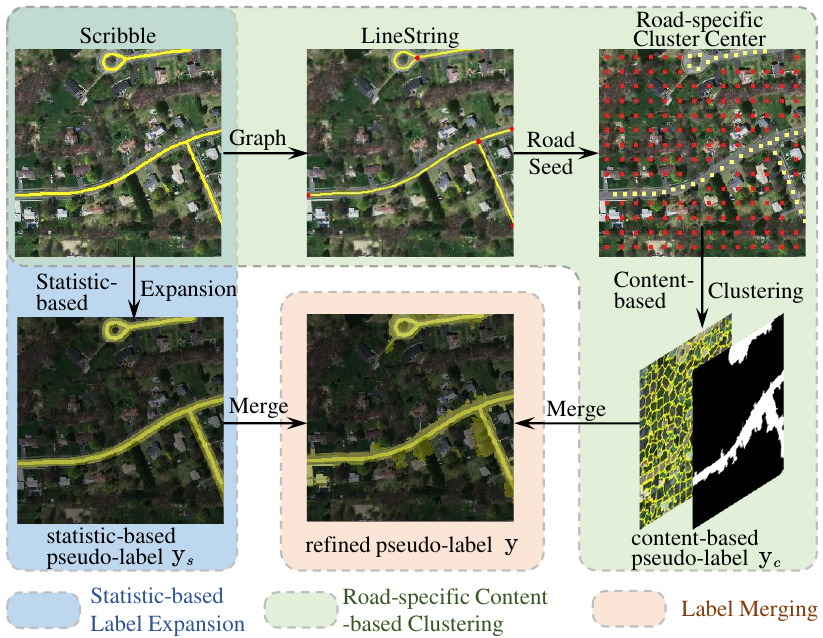} 
\caption{The flow chart of Statistic and Content-based Label Propagation, including statistic-based expansion (annotated with blue), content-based clustering (annotated with green), and the merge of statistic-based pseudo label $\mathbf{y}_s$ and content-based pseudo label $\mathbf{y}_c$ (annotated with orange).}
\label{RSG-Road} 
\end{figure}

The scribble annotation is easily accessible, but it rarely provides sufficient supervising information. The intuitive method is to expand the scribble annotation with width statistic information, but it may lead to poor adaptability on roads with various widths.  To improve the label expansion, we incorporate road-specific content-based clustering into statistic-based expansion, as illustrated in Fig.\ref{RSG-Road}.
% On the other hand, extracting pseudo labels conditioned on image content is likely to be affected by poor initial clustering centers. （这句话我想删掉，黄浩想加上，你核对下）

% To obtain more precise positioning of pseudo labels,  we incorporate road-specific content-based clustering into prior-based label expansion, as illustrated in Fig.\ref{RSG-Road}. 
% In content-based clustering, we take the road structure into consideration by initiating the clustering centers with road-specific seed points.

\paragraph{Statistic-based Label Expansion} 
According to the statistical knowledge of road width, $b_1$ and $b_2$ are chosen as the upper and lower limits of roads to perform the buffer expansion, respectively. 
The statistic-based pseudo labels $\mathbf{y}_s^i$ of the pixels $p_i$ is classified as
\begin{equation}
\label{buffer_zone}
\mathbf{y}_s^i=\left\{
\begin{aligned}
%\nonumber
1 \ \ \ (foreground) \ \ \ \ &\text{if} \ \ 0<\text{DIS}_{p_i}\leq b_1,\\
0.5 \ \ \ \ (uncertain) \ \ \ &\text{if} \ \ b_1<\text{DIS}_{p_i}\leq b_2,\\
0 \ \ \ (background) \ \ \ \ &\text{if} \ \ \text{DIS}_{p_i} > b_2,
\end{aligned}
\right.
\end{equation}
where $\text{DIS}_{p_i}$ denotes the distance from the pixel $p_i$ to the scribble. $\mathbf{y}_s^i\in\{\text{1},\text{0},\text{0.5}\} $ is the statistic-based label of the pixel $p_i$, which represents the class of \emph{foreground}, \emph{background}, and \emph{uncertain} region, respectively. 

\paragraph{Road-specific Content-based Clustering}
Clustering-based method \cite{2021Scribble, SLIC} improves the efficiency and explores similar pixels by grouping the pixels into meaningful regions. We further improve upon this method by using road-specific seed points as initial clustering centers, alleviating the issues caused by cluster's sensitivity to initial points and enhancing adherence to rough road boundaries.

% However, the superpixels generated by clustering often use uniform sampling as initial seed points and are difficult to adhere to rough road boundaries.
% To address this problem, a road-specific content-based clustering method is designed by selecting road-specific seed points that take into account the key points of road typology and the representative points of road connectivity.}

% For the scribble labels, the intersections, starting, and ending of roads are selected as the key points $d_{key}$. The representative points $d_{sp}$  are determined by sampling every $q$ point along the direction of each scribble.  The value of $q$ is related to the resolution of the image $(H\times W)$ and the number of superpixels $(N_{SLIC})$ in the whole image. $q$  is calculated as
For scribble labels, we select intersections and start/end points as key points $d_{key}$, and perform sampling with a stride of $q$  on each scribble to obtain representative points $d_{rep}$. Both $d_{key}$ and $d_{rep}$ collectively serve as foreground seeds. $q$ is calculated as
\begin{equation}
    q=\frac{H\cdot W}{\sqrt{2}\cdot\sqrt{N_{SLIC}}},
\end{equation}
where $N_{SLIC}$ is the number of superpixels and $(H\times W)$ is the resolution of the image. Background seeds are sampled at intervals of \(H \cdot W / N_{SLIC}\). Seeds within \((b_1 + b_2)/2\) pixels of the scribbles are discarded to prevent interference with road areas. We then use Graph Cut \cite{Cut_Loss} to categorize superpixels into \emph{background} and \emph{foreground} with potential road regions, generating content-based pseudo labels $\mathbf{y}_c^i\in\{0, 1\}$.
% \textcolor{blue}{The background seeds are uniformly sampled at  $H\cdot W/N_{SLIC}$ intervals, while the seeds within $(b_1+b_2)/2$   pixels from the scribbles are discarded to avoid interference for roads.  Subsequently, the Graph Cut\cite{Cut_Loss} is used to classify the superpixels into the background and potential road areas,  resulting in the content-based pseudo labels $\mathbf{y}_c^i\in\{0, 1\}$. }

\paragraph{Label Merging} 
% Error的内容
The statistic-based label expansion hardly provides the buffer 
areas that adapt to different roads, it may lead to a potential category conflict between $\mathbf{y}_c$ and $\mathbf{y}_s$.
Thus, the buffer setting is refined based on the content-based pseudo labels. The refined pseudo labels  $\bf{y}^i$  are defined as
\begin{equation}\label{eq:merging}
\mathbf{y}^{i}=\left\{
\begin{aligned}
%\nonumber
0.5 \ (uncertain) \ \ \ \ \  &\text{if}\  \mathbf{y}_s^i = 0\  \text{and} \ \mathbf{y}_c^i = 1,\\
\mathbf{y}_s^i \ \ \ \ \ \ \ \ \ \ \ \ \ \ \ &\text{otherwise}.
\end{aligned}
\right.
\end{equation}
In \cref{eq:merging},  if the pixel $p_i$ is identified as \emph{background} by statistic-based expansion and as \emph{foreground} by content-based expansion, it will be updated as \emph{uncertain} buffer. Otherwise, keep the statistic-based label  $\mathbf{y}_s^i$  unchanged.

%This step allows us to avoid being overly conservative in setting the upper limit of the buffer to prevent errors and enables us to obtain more supervisory information. 这句话也是我删掉，黄浩想留下，你核对下

\subsection{Structure-aware Mixup}

% Appropriate perturbation methods will provide effective inputs for the proposed CVC-Road, leading to improved performance and robustness of the model. 
% As analyzed in Section \ref{SC:Related Works}, existing perturbation methods have some shortcomings. To address these issues,

\begin{figure}[h] 
\centering
\includegraphics[width=0.48\textwidth]{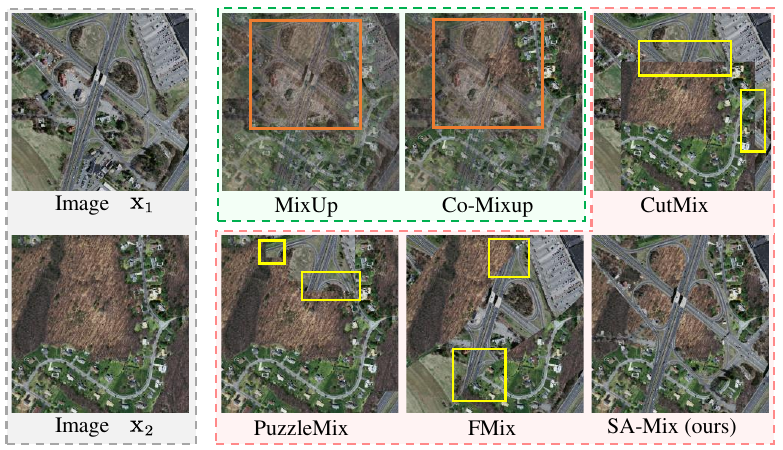} 
\caption{The visualization of different \emph{mixup} methods. Methods causing image overlay are green, and the indistinguishable regions are marked by the orange box; Non-overlay methods are red, and the damaged structures are marked by the yellow box. Our proposed SA-Mix has better road structure integrity compared to other methods, generating samples with proper scenes.}
\label{fig_mixup} 
\end{figure}

% , and restricting the consistency between original and constructed images is a promising learning strategy in weakly supervised learning.
Most existing mixup methods use block-based pasting while leaving the structural characteristics of the roads unconsidered, which thus risks generating impractical images, shown in \cref{fig_mixup}.
In order to construct samples with more complex image scenes, we propose a novel perturbation scheme to paste foreground regions with an explicit focus on roads' structure, namely structure-aware Mixup (SA-Mix). It constructs the images by significantly increasing the variety of road intersections and diverse combinations of foreground and background.

% \textcolor{blue}{By introducing suitable perturbations to input images, restricting the consistency between input and perturbed images is a promising learning strategy in weakly supervised learning. 
% Most existing mixup-based perturbation methods are based on block-based pasting while leaving the structural characteristics of the roads unconsidered, which thus risks generating impractical images, shown in Figure.\ref{fig_mixup}. 
% To enhance the topological structure of the roads, we propose a novel perturbation scheme to paste foreground regions with an explicit focus on roads' structure, namely structure-aware Mixup (SA-Mix). }

% When existing mixup methods are directly used for scribble-labeled road extraction, the uncertainty of road labels would cause cumulative bias. 
% Moreover, the sparsity distribution of roads may lead to more imbalances. 
% More seriously, existing region-based or generative-based mixup methods would cause the destruction of topology and connectivity of roads. 
% Thus, we propose a new structure-aware Mixup (SA-Mix) tailored to road extraction.

For a given pair of images ($\mathbf{x}_1, \mathbf{x}_2$), SA-Mix takes the refined pseudo label $(\mathbf{y}_1, \mathbf{y}_2)$ as a structural clue, and pastes the non-background regions onto the other image. The constructed images $(\mathbf{x}_{1 \leftarrow 2}^m, \mathbf{x}_{2 \leftarrow 1}^m)$ representing paste  $\mathbf{x}_2$ to $\mathbf{x}_1$ and $\mathbf{x}_1$ to $\mathbf{x}_2$ respectively are obtained as
\begin{align}
&\left\{
\begin{aligned}
\mathbf{x}^m_{1 \leftarrow 2}=&\text{D}_{\text{KL}}(\mathbf{x}_1,\mathbf{x}_2)\cdot[\mathbf{x}_1\odot(1-\mathbf{\alpha}_2)+\mathbf{x}_2\odot \mathbf{\alpha}_2]\\
&+[1-\text{D}_{\text{KL}}(\mathbf{x}_1,\mathbf{x}_2)]\cdot \mathbf{x}_1,\\
\mathbf{x}^m_{2 \leftarrow 1}=&\text{D}_{\text{KL}}(\mathbf{x}_1,\mathbf{x}_2)\cdot[\mathbf{x}_2\odot(1-\mathbf{\alpha}_1)+\mathbf{x}_1\odot \mathbf{\alpha}_1]\\
&+[1-\text{D}_{\text{KL}}(\mathbf{x}_1,\mathbf{x}_2)]\cdot \mathbf{x}_2,
\end{aligned}
\right.
\end{align}
% \begin{align}
% \nonumber
% &\text{SA-Mix}(\mathbf{x}_1,\mathbf{x}_2)=\mathbf{x}^m_{1 \to 2},\mathbf{x}^m_{2 \to 1}\\
% &\left\{
% \begin{aligned}
% \mathbf{x}^m_{1 \to 2}=&\text{D}_{\text{KL}}(\mathbf{x}_1,\mathbf{x}_2)\cdot[\mathbf{x}_1\odot(1-\mathbf{y}_c^2)+\mathbf{x}_2\odot \mathbf{y}_c^2]\\
% &+[1-\text{D}_{\text{KL}}(\mathbf{x}_1,\mathbf{x}_2)]\cdot \mathbf{x}_1\\
% \mathbf{x}^m_{2 \to 1}=&\text{D}_{\text{KL}}(\mathbf{x}_1,\mathbf{x}_2)\cdot[\mathbf{x}_2\odot(1-\mathbf{y}_c^1)+\mathbf{x}_1\odot \mathbf{y}_c^1]\\
% &+[1-\text{D}_{\text{KL}}(\mathbf{x}_1,\mathbf{x}_2)]\cdot \mathbf{x}_2
% \end{aligned}
% \right.
% \label{eq:SA-Mix}
% \end{align}
where $\odot$ refers to the element-wise multiplication, and $\mathbf{\alpha}_i=\mathbf{I}(\mathbf{y}_i>0)$ is the mask of non-background regions in $\mathbf{y}_i$ generated by the indicator function $\mathbf{I}(\cdot)$.

For preventing interfering edges on a pair of images with significant color differences,
we introduce a binary indicator $\text{D}_{\text{KL}}(\mathbf{x}_1,\mathbf{x}_2)$ based on the color similarity between the image pair in HSV space, defined as
\begin{equation}
\text{D}_{\text{KL}}(\mathbf{x}_1,\mathbf{x}_2)=\left\{
\begin{aligned}
&1 \ \ \ \text{if}\ \ \mathbf{KLDiv}(Hist_1,Hist_2)<t,\\
&0 \ \ \ \ \ \ \ \ \  \text{otherwise},
\end{aligned}
\right.\\
\end{equation}
where $\mathbf{KLDiv}(\cdot, \cdot)$ denotes the Kullback–Leibler (KL) divergence.
$t$ is a threshold for filtering out the image pairs with extreme color differences, such that interfering edges can be properly suppressed. 

For the filtered image pair, a similar mixing operator is applied on their pseudo labels, and the pseudo labels for the constructed images ($\mathbf{y}_{1 \leftarrow 2}^m, \mathbf{y}_{2 \leftarrow 1}^m$) are obtained as
\begin{equation}
\begin{aligned}
\mathbf{y}^m_{1 \leftarrow 2}=\mathbf{y}_1\odot(1-\mathbf{\alpha}_2)+\mathbf{y}_2\odot \mathbf{\alpha}_2,\\
\mathbf{y}^m_{2 \leftarrow 1}=\mathbf{y}_2\odot(1-\mathbf{\alpha}_1)+\mathbf{y}_1\odot \mathbf{\alpha}_1.
\end{aligned}
\end{equation}

\subsection{Supervision of SA-MixNet}
In our SA-MixNet, the training process is jointly optimized by segmentation loss, invariance regularization, and connectivity regularization. Along with such basic pseudo segmentation supervision, we propose invariance regularization$(\mathcal{L}_{inv})$ to force the model to approach consistent performance on original and constructed samples, enhancing the model's invariance under different conditions.
Moreover, discriminator-based connectivity regularization $(\mathcal{L}_{C-D})$ is designed to improve the topology connectivity of roads.
The overall loss $\mathcal{L}$ can be summarized as
\begin{equation}\label{eq:Loss}
\mathcal{L}=\mathcal{L}_{seg}+\mathcal{L}_{seg_\text{m}}+\lambda_1\mathcal{L}_{inv}+\lambda_2\mathcal{L}_{C-D},
\end{equation}
where, the contribution of these loss terms to the overall training loss is controlled by their corresponding coefficients $\lambda_1,\lambda_2$.

 $\mathcal{L}_{seg}$ and $\mathcal{L}_{seg_\text{m}}$  are  the segmentation losses designed to minimize the differences between predictions$(\mathbf{p})$ and pseudo labels$(\mathbf{y})$ on both original and constructed samples, defined as
\begin{align}
\begin{aligned}\label{eq:segloss}
&\mathcal{L}_{seg}=\frac{1}{2}[\mathcal{L}_{\text{p}}(\mathbf{y}_1, \mathbf{p}_1)+\mathcal{L}_{\text{p}}(\mathbf{y}_2, \mathbf{p}_2)],\\
&\mathcal{L}_{seg_\text{m}}=\frac{1}{2}[\mathcal{L}_{\text{p}}(\mathbf{y}^m_{1 \leftarrow 2},\mathbf{p}^m_{1 \leftarrow 2})+\mathcal{L}_{\text{p}}(\mathbf{y}^m_{2 \leftarrow 1},\mathbf{p}^m_{2 \leftarrow 1})],
\end{aligned}
\end{align}
where $\mathcal{L}_{\text{p}}(\cdot,\cdot)$ represents Partial BCE loss that measures Cross-Entropy on foreground and background pixels while ignoring those \textit{uncertain} ones.

\paragraph{Invariance Regularization}
Actually, the samples constructed by SA-Mixup are more challenging than the original ones by adding extra road intersections and combinations of backgrounds and foregrounds.
In order to ensure that the model has sustained outstanding performance in dealing with such challenging samples, directed invariance regularization is designed.
It is inspired by consistency learning, and it improves the model's learning ability for difficult samples by forcing the model to perform as well on the constructed samples as the original ones.  

% Since the constructed images contain the same foreground pixels as the original ones, the only difference is the various scenes contain the background. 
The expected invariance between predictions of constructed samples $\mathbf{p}^m$ and mixed original predictions $\mathbf{\overline{p}}^m$ are formulated as 
\begin{align}
\begin{aligned}\label{relation}
\mathcal{S}(\text{SA-Mix}(\mathbf{x}_1,\mathbf{x}_2))&=\text{SA-Mix}(\mathcal{S}(\mathbf{x}_1),\mathcal{S}(\mathbf{x}_2)),\\
\mathbf{p}_{1 \leftarrow 2}^m, \mathbf{p}_{2 \leftarrow 1}^m &= \mathbf{\overline{p}}_{1 \leftarrow 2}^m, \mathbf{\overline{p}}_{2 \leftarrow 1}^m,
\end{aligned}
\end{align}
where $\mathcal{S}(\cdot)$ represents the segment method, and $\mathcal{L}_{inv}$ is defined based on cosine similarity $\mathcal{L}_\text{cos}$ as
\begin{align}
\begin{aligned}\label{relation}
 &\mathcal{L}_{inv}=\frac{1}{2}[\mathcal{L}_\text{cos}(\mathbf{p}_{1 \leftarrow 2}^m, _\perp\mathbf{\overline{p}}_{1 \leftarrow 2}^m) +\mathcal{L}_\text{cos}(\mathbf{p}_{2 \leftarrow 1}^m,_\perp\mathbf{\overline{p}}_{2 \leftarrow 1}^m)],\\
&\mathcal{L}_\text{cos}(\mathbf{p}^m,_\perp\mathbf{\overline{p}}^m)=1-\frac{\mathbf{p}^m\cdot_\perp\mathbf{\overline{p}}^m}{\|\mathbf{p}^m\|\cdot\|_\perp\mathbf{\overline{p}}^m\|},
\end{aligned}
\end{align}
where $\perp$ is an operator that sets the gradient of the operand to zero. Consequently, the gradient of $\mathcal{L}_{inv}$ is truncated at the prediction of the original image and directed backpropagated to the constructed image.

\paragraph{Connectivity Regularization}
% In addition to the preservation of the road structure by SA-Mixup with Invariance Regularization, we further focus on the road's topological connectivity.
The connectivity regularization is designed to enhance the connectivity integrity of roads through adversarial learning without introducing additional supervising information. It is applied to original images' predictions.

\begin{figure}[h] 
\centering
\includegraphics[width=0.49\textwidth]{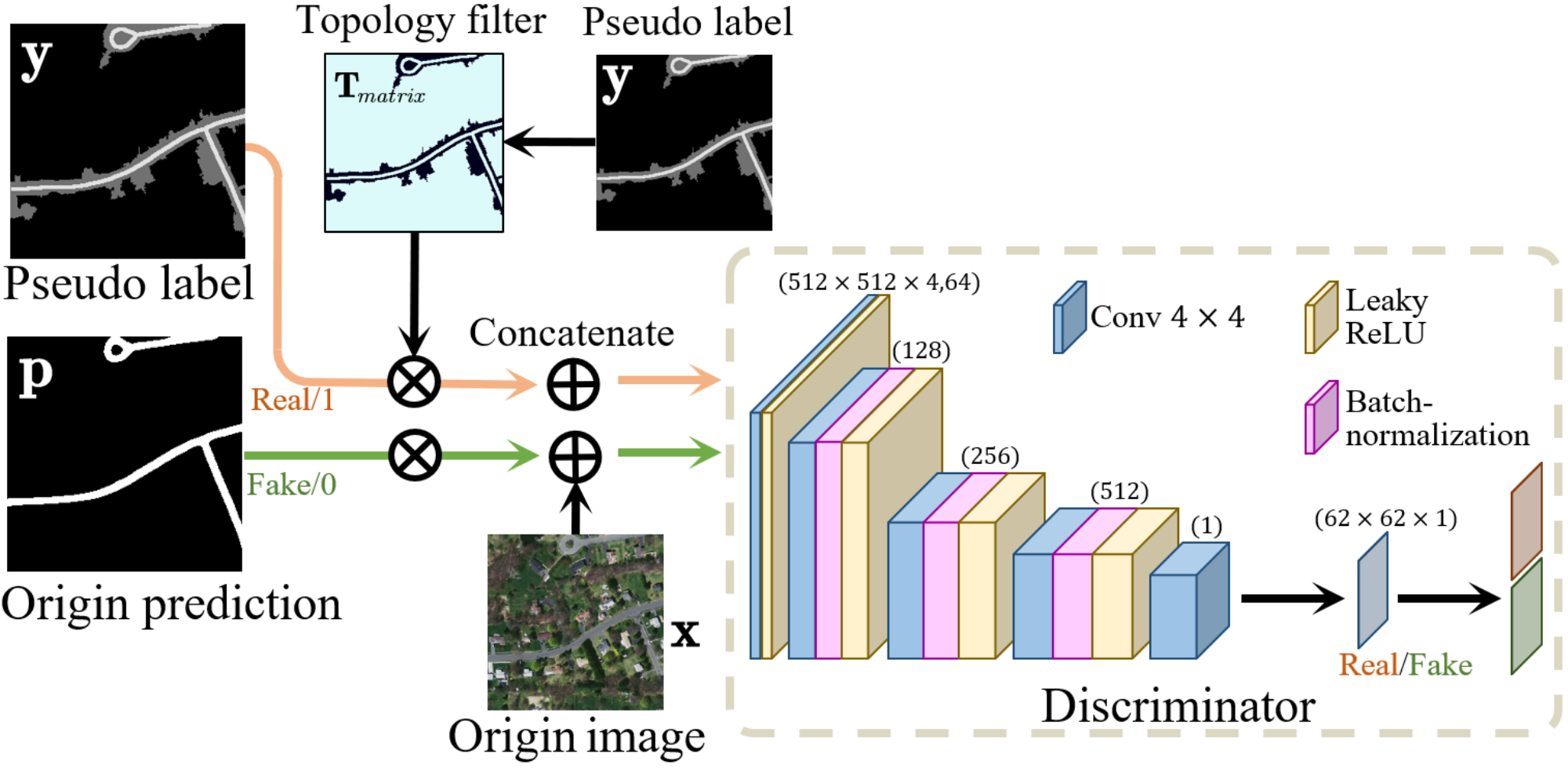} 
\caption{The pipeline of topological connectivity discriminator. The \emph{Pseudo label} $\mathbf{y}$ and \emph{Prediction} $\mathbf{p}$ are filtered by the \emph{Topology filter} $\mathbf{T}_{matrix}$ generated by the \emph{Pseudo label} $\mathbf{y}$, then concatenated with \emph{Image} $\mathbf{x}$ respectively, and input into the \emph{Discriminator}.}
\label{fig_Dis} 
\end{figure}

Utilizing the encoder-decoder module as the generator, we incorporate a topological connectivity network inspired by patch GAN \cite{PatchGAN} as the discriminator. The detailed architecture  is shown in the \cref{fig_Dis}

 First, we generate a topological filter matrix $\mathbf{T}_{matrix}$ based on $\mathbf{y}$, which is used to extract topological connectivity features and determined background features ignoring edge information in the \emph{uncertain} region of $\mathbf{p}$ and $\mathbf{y}$. The process is defined as
\begin{equation}
    \mathbf{p}_T, \mathbf{y}_T=\mathbf{p}\odot \mathbf{T}_{matrix}, \ \mathbf{y}\odot \mathbf{T}_{matrix},
\end{equation}
where
\begin{align}
\mathbf{T}_{matrix}=\left\{
\begin{aligned}
&1 \ \ \ \ \ \text{if}\ \ \mathbf{y}=1 \ \ \text{or} \ \ \mathbf{y}=0,\\
&0 \ \ \ \ \ \text{if}\ \ \mathbf{y}=0.5.
\end{aligned}
\right.
\end{align}
 
% \begin{align}\label{eq:matrix}
% (\mathbf{p}^T, \mathbf{y}^T)&=(\mathbf{p}, \mathbf{y})\odot \mathbf{T}_{matrix}\\
% \mathbf{T}_{matrix}&=\left\{
% \begin{aligned}
% &1 \ \ \ \ \ \text{if}\ \ \mathbf{y}=1 \ \ \text{or} \ \ \mathbf{y}=0\\
% &0 \ \ \ \ \ \text{if}\ \ \mathbf{y}=0.5
% \end{aligned}
% \right.
% \end{align}

\begin{table*}[htbp]
  \centering
  \begin{center}
    \resizebox{1.0\textwidth}{!}{
    \begin{tabular}{c|l|c@{\hspace{2.5mm}}c@{\hspace{2.5mm}}c@{\hspace{2.5mm}}c|c@{\hspace{2.5mm}}c@{\hspace{2.5mm}}c@{\hspace{2.5mm}}c|c@{\hspace{2.5mm}}c@{\hspace{2.5mm}}c@{\hspace{2.5mm}}c}
    \toprule
    \multicolumn{1}{c}{} & \multicolumn{1}{c|}{} & \multicolumn{4}{c|}{DeepGlobe} & \multicolumn{4}{c|}{Wuhan}     & \multicolumn{4}{c}{Massachusetts-road} \\
\cmidrule{1-14}    \multicolumn{1}{c|}{Method} & \multicolumn{1}{c|}{Description} & IoU   & F1    & Precision & Recall & IoU   & F1    & Precision & Recall & IoU   & F1    & Precision & Recall \\
    \midrule
    ScribbleSup\cite{2016ScribbleSup} & VGG-16  & 26.94  & 40.79  & 29.51  & \textbf{88.13} & 47.40  & 62.22  & 60.86  & \textbf{72.67} & ——    & ——    & ——    & —— \\
    WSOD\cite{WSOD}  & VGG-16  & 46.87  & 63.82  & 59.96  & 68.21  & 48.87  & 65.66  & 64.09  & 67.31  & 42.04  & 59.20  & 62.96  & 55.85  \\
    BPG\cite{BPG}  & DeepLabV2-101 & 54.00 & 70.13 & 66.90 &  73.68 & 50.86  & 67.43  & 76.66  & 60.19  & 55.45  & 71.34  & 66.32  & 77.18  \\
    \midrule
    MixUp\cite{2017_mixup} &  & 51.02 &  67.57 &  55.52 &  86.31 & 51.97 & 68.39 & \textbf{79.54} & 59.99 & 60.58 & 75.45 & 70.50 & 81.14 \\
    Co-Mixup\cite{Co-Mixup} & + D-LinkNet-34 & 52.78  & 69.10  & \textbf{85.08} & 58.18  & 52.71  & 69.03  & 78.68 & 61.49  & 60.67  & 75.52  & \textbf{75.63} & 75.42  \\
    FMix\cite{FMix}  & + proposed $\mathbf{y}$ & 57.57  & 73.07  & 75.04  & 71.21  & 53.73  & 69.90  & 75.00  & 65.45  & 60.81  & 72.63  & 72.96  & 78.50  \\
    CutMix\cite{2019_cutmix} & + proposed $\mathcal{L}$ & 58.28  & 73.64  & 74.73  & 72.59  & 52.93  & 69.22  & 70.89  & 67.63  & 61.45  & 76.12  & 73.30  & 79.18  \\
    PuzzleMix\cite{2020_puzzlemix} &  & 58.79  & 74.05  & 72.05  & 76.16  & 54.87  & 70.86  & 75.36  & 66.86  & 62.25  & 76.73  & 74.88  & 78.68  \\
    \midrule
    Baseline$_{\text{weak}}$ & D-LinkNet-34 & 57.00  & 72.61  & 72.88  & 72.35  & 52.56  & 68.91  & 67.65  & 70.22  & 55.34  & 71.25  & 62.43  & 82.98  \\
    WeaklyOSM\cite{2017Weakly} & MD-ResUnet & 54.32  & 70.40  & 68.85  & 72.03  & 52.34  & 68.72  & 67.93  & 69.52  & 52.74  & 69.06  & 57.83  & \textbf{85.67} \\
    ScRoad\cite{2019RoadScribble} & DBNet-34 & 58.91  & 74.14  & 70.79  & 77.82  & 53.64  & 69.82  & 75.29  & 65.10  & 58.19  & 73.57  & 67.89  & 80.28  \\
    SA-MixNet (ours)  & D-LinkNet-34 &\textbf{60.38} & \textbf{75.29} & 73.30  & 77.40  & \textbf{55.76} & \textbf{71.59} & 74.06  & 69.29  & \textbf{62.28} & \textbf{76.77} & 73.66  & 80.13  \\
    \bottomrule
    \end{tabular}
    }
  \end{center}
     \caption{Comparison of the proposed method against state-of-the-art generic, road-specific weakly supervised techniques and various strong mixup methods incorporating the proposed SA-MixNet framework on the DeepGlobe, Wuhan, and Massachusetts-road datasets.}
  \label{tab:table1}%
\end{table*}%

Then $\mathbf{p}_T$ and $\mathbf{y}_T$ are concatenated with images $\mathbf{x}$ to get $\mathbf{x}_d$ as input of the discriminator $(\mathcal{D})$. $\mathcal{D}$ outputs an evaluation vector of size $N\times N\times1$ to assess the topological integrity of roads and less false detection in each region. The cross-entropy loss is calculated for the true and false classes as 
\begin{align}
\label{eq:TD}
\begin{aligned}
\mathcal{L}_{C-D}=-&\sum_{h,w}(1-y_n)\text{log}(\mathcal{D}(\mathbf{x}_d)^{(h,w,0)})\\
&+y_n\text{log}(\mathcal{D}(\mathbf{x}_d)^{(h,w,1)}).
\end{aligned}
\end{align}
It forces the model to make predictions with less false detection and better connectivity.

\section{Experiment and Results}

\subsection{Datasets and Implementation}
\paragraph{Datasets}
We evaluate the proposed SA-MixNet on three challenging datasets, \textit{i.e.} DeepGlobe, Massachusetts-road, and Wuhan. 
These datasets cover a variety of scenes, including urban, rural and suburban areas.

DeepGlobe \cite{DeepGlobe_dataset} is one of the largest datasets of road extraction. 
It contains 6226 images in size of $1024\times1024$, and the ground resolution of each image is 50cm/pixel.
Following \cite{2021SPIN, 2019Improved}, the entire dataset is split into 4696 and 1530 images for training and testing, respectively.

\begin{figure}[t]
\centering
\includegraphics[width=0.48\textwidth]{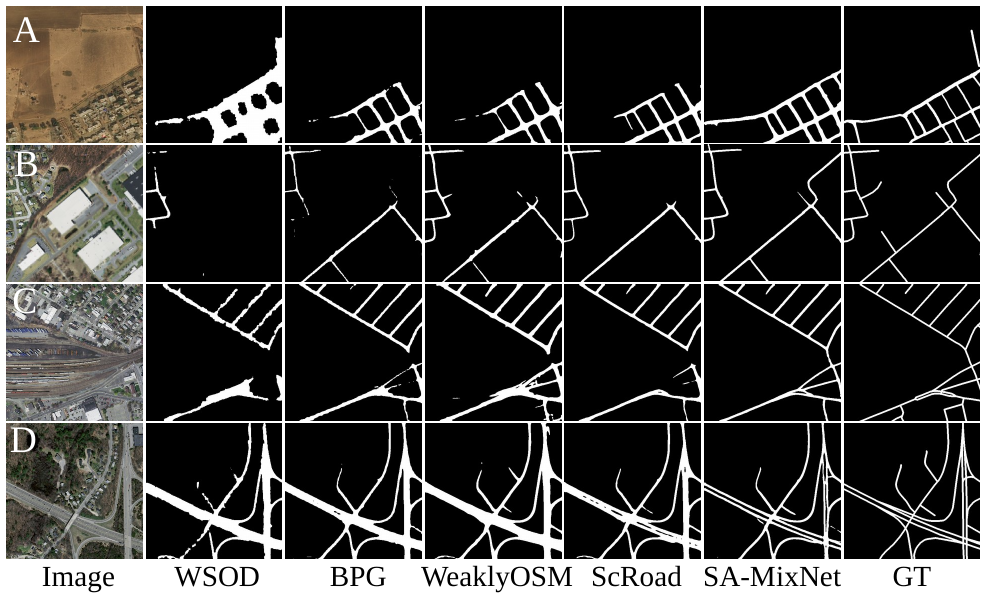} 
\caption{Visualization of GT, and WSOD, BPG, WeaklyOSM, ScRoadExtractor, SA-MixNet(ours)'s predictions of selected images with complex scenes. A) roads sharing similar features with background, B) blurry road areas, C) railways resembling road features, and D) overpasses with complex intersections}
\label{fig:vis_main}
\end{figure}

Massachusetts-road \cite{Ma_dataset} contains 1108 images, 14 images, and 49 images for training, validation, and testing.
These images are with a pixel resolution of $1500\times1500$, their ground resolution is 1.2m/pixel.
The road widths in this dataset are mostly consistent.

Wuhan is another dataset for road extraction containing 2592 images in size of $1024\times1024$ and a ground resolution of 50cm/pixel. 
The road widths in this dataset vary significantly. 
Following \cite{2021Scribble}, we create two splits with 1944 images and 648 images for training and testing, respectively.

While these datasets contain fully annotated road masks, the scribble labels are generated by skeletonizing the road masks.
Non-overlapping cropping is performed on DeepGlobe and Wuhan datasets, resulting in 18784 training and 6120 testing samples of $512\times512$ for DeepGlobe dataset, 7776 training and 2592 testing samples of $512\times512$ for Wuhan dataset. 
For Massachusetts-road, overlapping cropping is performed to get 9972, 126, and 441 samples for training, validation, and testing, respectively.

\begin{figure*}[h]
\centering
\includegraphics[width=0.9\textwidth]{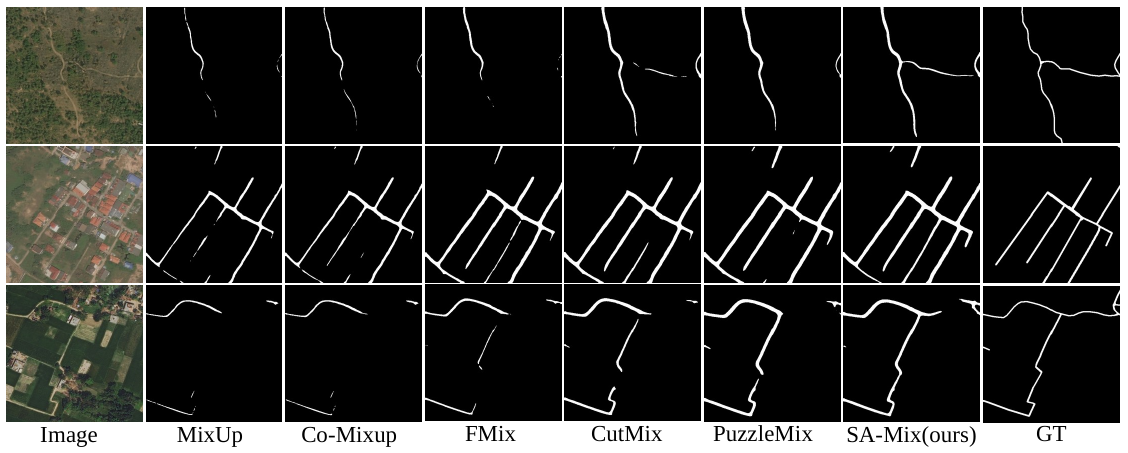} 
\caption{Visualization of MixUp, Co-Mixup, FMix, CutMix, PuzzleMix, and our SA-Mix's predictions. All mentioned \emph{mixup} methods use pseudo labels generated by our proposed SC-label expansion, and are supervised by proposed invariance-based regularization $\mathcal{L}$.}
\label{fig:vis_mix}
\end{figure*}

\paragraph{Implementation Details}
Our SA-MixNet is based on D-LinkNet \cite{2018D} with a ResNet-34 backbone. 
We employ the Adam optimizer to refine both the D-LinkNet and the discriminator. 
The initial learning rates are set to $2 \times 10^{-4}$ for D-LinkNet and halved for the discriminator.
The learning rates are decayed with a rate of 0.2 when the loss does not decrease for 6 epochs.
Early stopping is triggered when the loss remains unchanged for 12 epochs. 
During training, image flipping, rotation, mirroring and color shifts are adopted for data augmentation. 
Our model is trained with a batch size of 16.
For pseudo-label propagation, we adopt the same buffer zone setting as \cite{2021Scribble}.

% \noindent\textbf{Training Details} We utilize D-LinkNet based on ResNet34 as the encoder-decoder network for CVC-Road. Training data consists of images cropped to a resolution of $512\times512$, coupled with standard data augmentation techniques, including image flipping, rotation, mirroring, and color shifts. We employ the Adam optimizer to refine both the D-LinkNet and the discriminator with a batch size of 32. The learning rates are set to 2e-4 for D-LinkNet and 1e-4 for the discriminator and are divided by five if the loss does not decrease for six epochs. Early stopping is executed when the loss remains stagnant for 12 epochs or the learning rate drops below 5e-9. The proposed method is built on Pytorch and executed on Nvidia RTX3090 GPU with 24G of memory.
\subsection{Main Results}

\paragraph{Comparison with Weakly Supervised Methods}
\cref{tab:table1} summarizes the evaluation metrics by the proposed SA-MixNet on DeepGlobe, Wuhan and Massachusetds-Road datasets.

Compared against weakly supervised methods via scribble labels, the proposed SA-MixNet achieves the highest scores on these datasets in terms of IoU and F1 scores.
It achieves IoU gains of 1.47\%, 2.12\% and 4.09\% against current best performer ScRoadExtractor with an edge detection branch supervised by additional boundaries data on the DeepGlobe, Wuhan and Massachusetts-road datasets, respectively.
Surprisingly, SA-MixNet with the basic ResNet-34 backbone outperforms BPG with stronger DeepLabV2-101 backbone.

% We separately compare it with generic and road-specific weakly supervised methods via scribble labels: the iterative ScribbleSup\cite{2016ScribbleSup}, the non-iterative WSOD\cite{WSOD}, and BPG\cite{BPG} which incorporates an edge branch, the pioneering WeaklyOSM\cite{2019RoadScribble} and the current SOTA method, ScRoadExtractor\cite{2021Scribble}. 
% As shown in Table.\ref{tab:table1}, the proposed SA-MixNet achieves the best IOU and F1-score in three datasets. Compared to the current SOTA methods ScRoadExtractor, the proposed SA-MixNet achieved an IOU improvement of 0.65\%, 2.12\%, and 4.09\%, respectively, on the DeepGlobe, Wuhan, and Massachusetts-road datasets, and an F1-score improvement of 0.52\%, 1.77\%, and 3.2\%.
% In addition, the proposed method outperformed the benchmark generic weakly supervised methods using stronger backbones (DeepLab, ResNet101) on all datasets. 

For qualitative evaluation, a set of complex scenes are selected, including roads sharing similar features with background, blurry road areas, railways resembling road features, and overpasses with complex intersections. 
As visualized in \cref{fig:vis_main}, our SA-MixNet outperforms other methods in adapting to diverse scenarios. 
It delivers better edge and road structure than BPG and ScRoadExtractor where an extra boundary detection branch is deployed, even under conditions of similar features, unclear edges and pixel blurriness. 
Such improvement demonstrates that our SA-MixNet is capable of extracting road features with remarkable invariance across diverse scenes.
Moreover, the road masks predicted by our SA-MixNet also exhibit better road connectivity than other methods.
This may partly be attributed to the connectivity regularization during model training, and further analyses will be conducted in subsequent \cref{sc:ablation}.

% We visualize the predicted road maps from WSOD, BPG, WeaklyOSM, ScRoadExtractor, and our proposed SA-MixNet. Shown in Fig.\ref{fig:vis_main}. 
% The prediction results of SA-MixNet have fewer false positives and false negatives than the general weakly supervised algorithm.
% Furthermore, compared to BPG and ScRoadExtractor, which have independent edge detection branches, our method predicted clearer road edges. 
% A similar situation also occurred in WeaklyOSM regularized by Cut Loss.
% This validates the effectiveness of our designed consistency learning.
% Our results also show superior road connectivity, it's may due to the SA-Mix designed to preserve the road structure's integrity and the topological structure loss $\mathcal{L}_{T-D}$. Further analyses will be conducted in subsequent experiments.

\paragraph{Comparison with Mix-up Method}

To verify the effectiveness of our proposed SA-Mix module, we replaced it with various strong mixup schemes \cite{2017_mixup, Co-Mixup, FMix, 2019_cutmix, 2020_puzzlemix}.
For a fair comparison, the identical segmentor and pseudo-label generation strategy to our SA-MixNet are adopted for these methods, as well as the same learning framework.
% The pseudo-labels $\mathbf{y}$ used for training and the proposed loss $\mathcal{L}$, are identical to those in SA-MixNet.

As summarized in \cref{tab:table1} the proposed SA-Mix outperforms the best-listed PuzzleMix method by 1.59\%, 0.89\% on DeepGlobe and Wuhan dataset in terms of the IoU metric.
Remote sensing imagery encapsulates a higher level of complexity and intricacy compared to standard natural images. 
Such complexity renders techniques like Mixup and Co-Mixup less effective due to issues of overlapping.
As visualized in \cref{fig:vis_mix}, Mixup and Co-Mixup achieve inferior predictive performance, as the indistinguishability caused by overlapping impedes their ability to accurately capture features. 
The predicted road masks by non-heuristic methods, FMix and CutMix, suffer from severe fragmentation because the topological structure of roads is not considered. 
The heuristic PuzzleMix method is capable of preserving road structures to a certain extent during the patching process.
Unlike the above mixup schemes, our proposed SA-Mix method achieves the best performance in maintaining topological integrity, signifying its superior capability in preserving the continuity of road topology.

\subsection{Ablation Study}
\label{sc:ablation}
In this section, we examine the contribution of segmentation loss, invariance regularization and connectivity regularization to the superior performance improvement by our SA-MixNet.

% The results of ablation on the DeepGlobe dataset are shown in \cref{tab:ablation}.

% \begin{table}[h]
% \centering
% \footnotesize
% \caption{Ablation Study}
% \begin{tabular}{cccc|cccc}
% \toprule
%  \multicolumn{4}{c|}{Loss} & \multicolumn{4}{c}{DeepGlobe} \\
% \midrule
%   $\mathcal{L}_{\text{seg}}$ & $\mathcal{L}_{\text{seg}_\text{m}}$ & $\mathcal{L}_{\text{con}}$ & $\mathcal{L}_{T-D}$ & IOU & F1 & Precision & Recall\\
%   \midrule
%   \checkmark & & & & 58.81 & 74.07 & 75.17 & 73.00 \\
%   \checkmark & \checkmark & & & 59.34 & 74.48 & 73.48 & 74.48 \\
%   \checkmark & & \checkmark & & 59.25 & 74.41 & 72.96 & 75.93 \\
%   \checkmark & \checkmark & \checkmark & & 60.09 & 75.07 & 72.43 & 77.91 \\
%   \checkmark & \checkmark & \checkmark & \checkmark & 60.38 & 75.29 & 73.30 & 77.40 \\
% \bottomrule
% \end{tabular}
% \label{tab:ablation}
% \end{table}

\begin{table}[h]
\centering
\footnotesize
\begin{tabular}{c@{\hspace{1mm}}c@{\hspace{1mm}}c@{\hspace{1mm}}c|cccc}
\toprule
 \multicolumn{4}{c|}{Loss} & \multicolumn{4}{c}{DeepGlobe} \\
\midrule
  $\mathcal{L}_{\text{seg}}$ & $\mathcal{L}_{\text{seg}_\text{m}}$ & $\mathcal{L}_{inv}$ & $\mathcal{L}_{C-D}$ & IoU & F1 & Precision & Recall\\
  \midrule
  \checkmark & & & & 58.81 & 74.07 & 75.17 & 73.00 \\
  \checkmark & \checkmark & & & 59.34 & 74.48 & 73.48 & 74.48 \\
  % \checkmark & & \checkmark & & 59.25 & 74.41 & 72.96 & 75.93 \\
  \checkmark & \checkmark & \checkmark & & 60.09 & 75.07 & 72.43 & 77.91 \\
  \checkmark & \checkmark & \checkmark & \checkmark & 60.38 & 75.29 & 73.30 & 77.40 \\
\bottomrule
\end{tabular}
\caption{Ablation study of loss design on DeepGlobe dataset}
\label{tab:ablation}
\end{table}

% \noindent\textbf{Pesudo-label generation.}
% When D-LinkNet34 with binary cross entropy loss$(\mathcal{L}_{seg})$ using pseudo-labels generated by our PC-Label Expansion, the IOU performance increased by $0.14\%, 0.33\%, 1.14\%$ respectively on three datasets than using ScLabel\cite{2021Scribble}. This shows that PCLE generates more accurate pseudo-labels and proves its effectiveness.

% \noindent\textbf{Loss design.}

Our baseline model only adopts the segmentation loss $\mathcal{L}_{seg}$ for supervision.
As summarized in \cref{tab:ablation}, the introduction of SA-Mixup for sample construction, supervised by $\mathcal{L}_{seg_\text{m}}$, leads to an IoU gain of 0.53\%. 
This highlights the positive impact of a more diverse and challenging set of learning samples on model efficacy.

Additionally, activating the proposed invariance regularization yields a further IoU improvement of 0.75\%. 
This validates that, by forcing the model to align its performance on constructed samples with that on original ones, this invariance regularization helps identify ambiguous samples. 
Along with the improvement on IoU scores, the recall rate is also increased by adopting $\mathcal{L}_{seg_\text{m}}$ and $\mathcal{L}_{inv}$.
This validates that $\mathcal{L}_{seg_\text{m}}$ and $\mathcal{L}_{inv}$ help enhance the model's ability to recognize road features. 
With a significant increase of 3.43\% in recall rates, $\mathcal{L}_{inv}$ is believed to play a more important role in enhancing the model's invariance across various scenes.

Furthermore, by introducing $\mathcal{L}_{C-D}$, an additional 0.29\% improvement in IoU score is achieved. This highlights its role in reducing false positives and enhancing connectivity.

% When the components of mix-based regularization, $\mathcal{L}_{seg_\text{m}}$ and $\mathcal{L}_{con}$, are applied separately, there are significant$(0.38\%,1.35\%,0.86\%)$ and slight$(0.15\%,1.02\%,0.6\%)$ IOU improvements. 
% This proves that the two losses are constrained in terms of the accuracy and stability of the model, respectively 
% When applied together as mix-based regularization, it achieves a more significant IOU improvement$(0.39\%,1.06\%,0.98\%)$, indicating that the proposed mix-based regularization enhances the model's performance with limited data.
% Additionally, the addition of topology preserving constraint $\mathcal{L}_{T-D}$ brings the IOU improvement by $0.36\%,1.55\%,0.46\%$, demonstrating its effectiveness to improve the structural integrity and background accuracy(lee false positive).

% \begin{figure}[h]
% \centering
% \includegraphics[width=0.48\textwidth]{figures/vis_ablation.png} 
% \caption{VIS ABLATION \textcolor{Green}{6*2, only two samples from DeepGlobe, with some descriptions and subscriptions}}
% \label{fig:vis_ablation}
% \end{figure}

\subsection{Data Sensitivity Study}

We investigated the annotation sensitivity of the proposed SA-MixNet with varying ratios of scribble to full annotations on DeepGlobe dataset.
% \begin{table}[htbp]
%   \centering
%   {\footnotesize
%     \begin{tabular}{
%     S[table-format=3.0, table-space-text-post={\%}] @{\,:\,}
%     S[table-format=3.0, table-space-text-post={\%}]
%     |S[table-format=2.2]
%     S[table-format=2.2]
%     S[table-format=2.2]
%     S[table-format=2.2]}
%     \toprule
%     \multicolumn{2}{c|}{Weak : Full} & {IOU} & {F1} & {Precision} & {Recall} \\
%     \midrule
%     %80\% & 0\% &  &  &  &  \\
%     100\% & 0\% & 60.38 & 75.29 & 73.30 & 77.40 \\
%     90\% & 10\% & 63.18 & 77.44 & 75.29 & 79.71 \\
%     75\% & 25\% & 65.87 & 79.42 & 78.12 & 80.76 \\
%     50\% & 50\% & 67.06 & 80.28 & 78.44 & 82.22 \\
%     25\% & 75\% & 67.69 & 80.73 & 79.81 & 81.68 \\
%     0\% & 100\% & 68.10 & 81.02 & 79.52 & 82.58 \\
%     \midrule % 绘制细横线
%     \multicolumn{2}{c|}{$\text{D-LinkNet}_{w}$} & 57.00 & 72.61 & 72.88 & 72.35 \\ % 合并前两列并添加新行1的内容
%     \multicolumn{2}{c|}{$\text{D-LinkNet}_{f}$} & 64.95 & 78.75 & 81.44 & 76.23 \\ % 合并前两列并添加新行2的内容
%     \bottomrule % 表格底部的粗横线
%     \end{tabular}%
%   }
%     \caption{Data Sensitive Study: performance with different ratio of weak to full annotations}
%   \label{tab:senstive}%
% \end{table}
\begin{table}[t]
  \centering
  {\footnotesize
    \begin{tabular}{
    S[table-format=3.0, table-space-text-post={\%}] @{\,:\,}
    S[table-format=3.0, table-space-text-post={\%}]
    |S[table-format=2.2]
    S[table-format=2.2]
    S[table-format=2.2]
    S[table-format=2.2]}
    \toprule
    \multicolumn{2}{c|}{Weak : Full} & {IoU} & {F1} & {Precision} & {Recall} \\
    \midrule
    %80\% & 0\% &  &  &  &  \\
    \text{100\%} & \text{0\%} & \text{60.38} & \text{75.29} & \text{73.30} & \text{77.40} \\
    \text{90\%} & \text{10\%} & \text{63.18} & \text{77.44} & \text{75.29} & \text{79.71} \\
    \text{75\%} & \text{25\%} & \text{65.87} & \text{79.42} & \text{78.12} & \text{80.76} \\
    \text{50\%} & \text{50\%} & \text{67.06} & \text{80.28} & \text{78.44} & \text{82.22} \\
    \text{25\%} & \text{75\%} & \text{67.69} & \text{80.73} & \text{79.81} & \text{81.68} \\
    \text{0\%} & \text{100\%} & \text{68.10} & \text{81.02} & \text{79.52} & \text{82.58} \\
    \midrule
    \multicolumn{2}{c|}{$\text{D-LinkNet}_{w}$} & \text{57.00} & \text{72.61} & \text{72.88} & \text{72.35} \\
    \multicolumn{2}{c|}{$\text{D-LinkNet}_{f}$} & \text{64.95} & \text{78.75} & \text{81.44} & \text{76.23} \\
    \bottomrule
    \end{tabular}%
  }
    \caption{Data sensitive study on DeepGlobe dataset}
  \label{tab:senstive}%
\end{table}

As shown in \cref{tab:senstive}, our SA-MixNet, with 100\% scribble annotations, surpasses the baseline (D-LinkNet) by 3.38\% in IoU. The IoU score gradually increases, as more full annotations are fed for training.
It tends stable after the portion of full annotation reaches 50\%. 
Noteworthily, our method's IoU outperforms the fully supervised benchmark with just 25\% full annotations, and, with complete annotations, it exceeds this benchmark by 3.15\%.
These results imply that SA-MixNet, with a few increase in full annotations, can markedly enhance performance at various annotation levels and it is also highly effective under fully supervised conditions.

% Results in \cref{tab:senstive} indicated that with 100\% scribble annotations, SA-MixNet outperformed the baseline (D-LinkNet) by 3.38\% in IOU. Notably, IOU gains increased with more full annotations, plateauing at 50\% full annotation. Remarkably, with just 25\% full annotations, our method's IOU already exceeded the fully supervised benchmark, and with full annotations used entirely, it surpassed the benchmark by 3.05\%. 

% These findings suggest that SA-MixNet requires a few additional full annotations to significantly boost performance across different levels of annotation. It can also be applied to significantly improve performance under fully supervised conditions

\subsection{Generalizability on Different Road Extractors}
\label{subs:gen}

To further validate the generalizability of our proposed framework, SA-MixNet, across different models, we replace the encoder-decoder network with ScRoadExtractor's dual-branch network (DBNet)\cite{2021Scribble} and compare the performance with origin DBNet on DeepGlobe dataset. The pseudo labels generated by our SC-label expansion are applied.

% Table generated by Excel2LaTeX from sheet 'trans1'
\begin{table}[h]
  \centering
  \begin{center}
    \resizebox{1.0\linewidth}{!}{
  {\footnotesize
    \begin{tabular}{c|cc|cc|cc}
    \toprule
     Method & \multicolumn{2}{c|}{$\text{DBNet}_{\text{w}/edge}$} & \multicolumn{2}{c|}{$\text{Ours}\ +\ \text{DBNet}_{\text{wo}/edge}$} & \multicolumn{2}{c}{$\text{Ours}\ +\ \text{DBNet}_{\text{w}/edge}$}\\
     \midrule
    dataset & IoU & F1 & IoU & F1 & IoU & F1 \\
    \midrule
    DeepGlobe & 59.48 & 74.14 & 60.56 & 75.43 & 60.50 & 75.39\\
    Wuhan   & 54.45 & 70.50 & 55.38 & 71.28 & 54.45 & 70.51 \\
    Ma    & 59.47 & 74.59 & 62.60 & 77.00 & 62.36 & 76.81 \\
    \bottomrule
    \end{tabular}%
}}
\end{center}
  \caption{Generalizability Experiment}
  \label{tab:generalize}%
\end{table}%

As shown in \cref{tab:generalize}, after integrating with the proposed framework to DBNet without the edge detection branch, the IoU metrics on three datasets improved by 1.08\%, 0.93\%, and 3.13\%, respectively. 
This demonstrates that SA-MixNet can serve as a universal framework to extract more robust and intrinsic features by enhancing model invariance in varied scenes without additional supervision, showing its ability to significantly improve the performance of existing methods under data-constrained conditions. When applying SA-MixNet to complete DBNet, the performance is degraded compared to the case without edge detection branch, demonstrating that the proposed data-driven method can replace the additional supervision brought by the prior.

\section{Conclusions}

In this paper, we propose SA-MixNet, a novel framework for scribble-based road extraction from remote sensing images from a data-driven perspective, eliminating the requirements for additional priors. The proposed SA-Mixup is an efficient scheme to construct samples with various complex scenes, and the invariance regularization significantly improves the generalization ability of the model by forcing it to behave consistent performance in complex and normal scenes and learning invariance feature of the targets.

% We reformulated the learning process as consistency learning, and took road structure into specific consideration in pseudo-label generation, sample perturbation, and consistency regularization, resulting in more adequate annotations and more sufficient regularization. 
% SA-MixNet was evaluated on three open datasets, achieved state-of-the-art performance, and showed its potential for plug-and-play.
{
    \small
    \bibliographystyle{ieeenat_fullname}
    \bibliography{main}

\begin{thebibliography}{35}
\providecommand{\natexlab}[1]{#1}
\providecommand{\url}[1]{\texttt{#1}}
\expandafter\ifx\csname urlstyle\endcsname\relax
  \providecommand{\doi}[1]{doi: #1}\else
  \providecommand{\doi}{doi: \begingroup \urlstyle{rm}\Url}\fi

\bibitem[Achanta et~al.(2012)Achanta, Shaji, Smith, Lucchi, Fua, and Süsstrunk]{SLIC}
R. Achanta, A. Shaji, K. Smith, A. Lucchi, P. Fua, and Sabine Süsstrunk.
\newblock Slic superpixels compared to state-of-the-art superpixel methods.
\newblock \emph{IEEE Transactions on Pattern Analysis and Machine Intelligence}, page 2274–2282, 2012.

\bibitem[Bandara et~al.(2022)Bandara, Valanarasu, and Patel]{2021SPIN}
Wele Gedara~Chaminda Bandara, Jeya Maria~Jose Valanarasu, and Vishal~M Patel.
\newblock Spin road mapper: Extracting roads from aerial images via spatial and interaction space graph reasoning for autonomous driving.
\newblock pages 343--350, 2022.

\bibitem[Batra et~al.(2019)Batra, Singh, Pang, Basu, and Paluri]{2019Improved}
A. Batra, S. Singh, G. Pang, S. Basu, and M. Paluri.
\newblock Improved road connectivity by joint learning of orientation and segmentation.
\newblock In \emph{CVPR}, 2019.

\bibitem[Demir et~al.(2018)Demir, Koperski, Lindenbaum, Pang, Huang, Basu, Hughes, Tuia, and Raskar]{DeepGlobe_dataset}
Ilke Demir, Krzysztof Koperski, David Lindenbaum, Guan Pang, Jing Huang, Saikat Basu, Forest Hughes, Devis Tuia, and Ramesh Raskar.
\newblock Deepglobe 2018: A challenge to parse the earth through satellite images.
\newblock In \emph{2018 IEEE/CVF Conference on Computer Vision and Pattern Recognition Workshops (CVPRW)}, 2018.

\bibitem[Harris et~al.(2020)Harris, Marcu, Painter, Niranjan, Pr{\"u}gel-Bennett, and Hare]{FMix}
Ethan Harris, Antonia Marcu, Matthew Painter, Mahesan Niranjan, Adam Pr{\"u}gel-Bennett, and Jonathon Hare.
\newblock Fmix: Enhancing mixed sample data augmentation.
\newblock \emph{arXiv preprint arXiv:2002.12047}, 2020.

\bibitem[Hong et~al.(2017)Hong, Kwak, and Han]{2017Weakly}
S. Hong, S. Kwak, and B. Han.
\newblock Weakly supervised learning with deep convolutional neural networks for semantic segmentation: Understanding semantic layout of images with minimum human supervision.
\newblock \emph{IEEE Signal Processing Magazine}, 34\penalty0 (6):\penalty0 39--49, 2017.

\bibitem[Kim et~al.(2020)Kim, Choo, and Song]{2020_puzzlemix}
Jang-Hyun Kim, Wonho Choo, and HyunOh Song.
\newblock Puzzle mix: Exploiting saliency and local statistics for optimal mixup.
\newblock \emph{International Conference on Machine Learning}, 2020.

\bibitem[Kim et~al.(2021)Kim, Choo, Jeong, and Song]{Co-Mixup}
Jang-Hyun Kim, Wonho Choo, Hosan Jeong, and Hyun~Oh Song.
\newblock Co-mixup: Saliency guided joint mixup with supermodular diversity.
\newblock \emph{arXiv preprint arXiv:2102.03065}, 2021.

\bibitem[Laine and Aila(2016)]{2016Temporal}
Samuli Laine and Timo Aila.
\newblock Temporal ensembling for semi-supervised learning.
\newblock \emph{arXiv preprint arXiv:1610.02242}, 2016.

\bibitem[Li et~al.(2020)Li, Wang, Zhang, Liu, Mei, and Li]{2020topology}
Xingang Li, Yuebin Wang, Liqiang Zhang, Suhong Liu, Jie Mei, and Yang Li.
\newblock Topology-enhanced urban road extraction via a geographic feature-enhanced network.
\newblock \emph{IEEE Transactions on Geoscience and Remote Sensing}, page 8819–8830, 2020.

\bibitem[Lian and Huang(2021)]{2021WeaklyPoint}
R. Lian and L. Huang.
\newblock Weakly supervised road segmentation in high-resolution remote sensing images using point annotations.
\newblock \emph{IEEE Transactions on Geoscience and Remote Sensing}, PP\penalty0 (99):\penalty0 1--13, 2021.

\bibitem[Lin et~al.(2016)Lin, Dai, Jia, He, and Sun]{2016ScribbleSup}
Di Lin, Jifeng Dai, Jiaya Jia, Kaiming He, and Jian Sun.
\newblock Scribblesup: Scribble-supervised convolutional networks for semantic segmentation.
\newblock pages 3159--3167, 2016.

\bibitem[Mei et~al.(2021)Mei, Li, Gao, and Cheng]{2021CoANet}
Jie Mei, Rou-Jing Li, Wang Gao, and Ming-Ming Cheng.
\newblock Coanet: Connectivity attention network for road extraction from satellite imagery.
\newblock \emph{IEEE Transactions on Image Processing}, page 8540–8552, 2021.

\bibitem[Mnih(2013)]{Ma_dataset}
Volodymyr Mnih.
\newblock Machine learning for aerial image labeling.
\newblock 2013.

\bibitem[Obukhov et~al.(2019)Obukhov, Georgoulis, Dai, and Van~Gool]{CRF_Loss}
Anton Obukhov, Stamatios Georgoulis, Dengxin Dai, and Luc Van~Gool.
\newblock Gated crf loss for weakly supervised semantic image segmentation.
\newblock \emph{arXiv preprint arXiv:1906.04651}, 2019.

\bibitem[Ouali et~al.(2020)Ouali, Hudelot, and Tami]{2020Semi}
Yassine Ouali, Celine Hudelot, and Myriam Tami.
\newblock Semi-supervised semantic segmentation with cross-consistency training.
\newblock In \emph{Proceedings of the IEEE/CVF Conference on Computer Vision and Pattern Recognition (CVPR)}, 2020.

\bibitem[Pan et~al.(2022)Pan, Zhu, Zhang, Cao, Wang, Zhang, Han, and Hu]{2022Learning}
Junwen Pan, Pengfei Zhu, Kaihua Zhang, Bing Cao, Yu Wang, Dingwen Zhang, Junwei Han, and Qinghua Hu.
\newblock Learning self-supervised low-rank network for single-stage weakly and semi-supervised semantic segmentation.
\newblock \emph{International Journal of Computer Vision}, 130\penalty0 (5):\penalty0 1181--1195, 2022.

\bibitem[Rasmus et~al.(2015)Rasmus, Berglund, Honkala, Valpola, and Raiko]{Rasmus_Valpola_Honkala_Berglund_Raiko_2015}
Antti Rasmus, Mathias Berglund, Mikko Honkala, Harri Valpola, and Tapani Raiko.
\newblock Semi-supervised learning with ladder networks.
\newblock \emph{Advances in neural information processing systems}, 28, 2015.

\bibitem[Sajjadi et~al.(2016)Sajjadi, Javanmardi, and Tasdizen]{2016Regularization}
Mehdi Sajjadi, Mehran Javanmardi, and Tolga Tasdizen.
\newblock Regularization with stochastic transformations and perturbations for deep semi-supervised learning, 2016.

\bibitem[Sinop and Grady(2007)]{Sinop_Grady_2007}
Ali~Kemal Sinop and Leo Grady.
\newblock A seeded image segmentation framework unifying graph cuts and random walker which yields a new algorithm.
\newblock In \emph{2007 IEEE 11th International Conference on Computer Vision}, 2007.

\bibitem[Tack et~al.(2022)Tack, Yu, Jeong, Kim, Hwang, and Shin]{Tack_Yu_Jeong_Kim_Hwang_Shin_2022}
Jihoon Tack, Sihyun Yu, Jongheon Jeong, Minseon Kim, Sung~Ju Hwang, and Jinwoo Shin.
\newblock Consistency regularization for adversarial robustness.
\newblock \emph{Proceedings of the AAAI Conference on Artificial Intelligence}, page 8414–8422, 2022.

\bibitem[Tang et~al.(2018)Tang, Djelouah, Perazzi, Boykov, and Schroers]{Cut_Loss}
Meng Tang, Abdelaziz Djelouah, Federico Perazzi, Yuri Boykov, and Christopher Schroers.
\newblock Normalized cut loss for weakly-supervised cnn segmentation.
\newblock In \emph{2018 IEEE/CVF Conference on Computer Vision and Pattern Recognition}, 2018.

\bibitem[Tao et~al.(2018)Tao, Chen, Yang, and Yin]{2018Stacked}
S. Tao, Z. Chen, W. Yang, and W. Yin.
\newblock Stacked u-nets with multi-output for road extraction.
\newblock In \emph{2018 IEEE/CVF Conference on Computer Vision and Pattern Recognition Workshops (CVPRW)}, 2018.

\bibitem[Wang et~al.(2019)Wang, Qi, Tang, Zhang, Wei, Li, and Zhang]{BPG}
Bin Wang, Guojun Qi, Sheng Tang, Tianzhu Zhang, Yunchao Wei, Linghui Li, and Yongdong Zhang.
\newblock Boundary perception guidance: A scribble-supervised semantic segmentation approach.
\newblock In \emph{Proceedings of the Twenty-Eighth International Joint Conference on Artificial Intelligence}, 2019.

\bibitem[Wei and Ji(2021)]{2021Scribble}
Y. Wei and S. Ji.
\newblock Scribble-based weakly supervised deep learning for road surface extraction from remote sensing images.
\newblock \emph{IEEE Transactions on Geoscience and Remote Sensing}, \penalty0 (99), 2021.

\bibitem[Wu et~al.(2019)Wu, Du, Chen, Xu, Guo, and Jing]{2019RoadScribble}
S. Wu, C. Du, H. Chen, Y. Xu, N. Guo, and N. Jing.
\newblock Road extraction from very high resolution images using weakly labeled openstreetmap centerline.
\newblock \emph{International Journal of Geo-Information}, \penalty0 (11), 2019.

\bibitem[Xu et~al.(2022)Xu, Zeng, Lian, and Ding]{Xu_Zeng_Lian_Ding_2022}
Bingrong Xu, Zhigang Zeng, Cheng Lian, and Zhengming Ding.
\newblock Generative mixup networks for zero-shot learning.
\newblock \emph{IEEE Transactions on Neural Networks and Learning Systems}, page 1–12, 2022.

\bibitem[Yun et~al.(2019)Yun, Han, Chun, Oh, Yoo, and Choe]{2019_cutmix}
Sangdoo Yun, Dongyoon Han, Sanghyuk Chun, Seong~Joon Oh, Youngjoon Yoo, and Junsuk Choe.
\newblock Cutmix: Regularization strategy to train strong classifiers with localizable features.
\newblock In \emph{2019 IEEE/CVF International Conference on Computer Vision (ICCV)}, 2019.

\bibitem[Zhang et~al.(2017)Zhang, Cisse, Dauphin, and Lopez-Paz]{2017_mixup}
Hongyi Zhang, Moustapha Cisse, Yann~N Dauphin, and David Lopez-Paz.
\newblock mixup: Beyond empirical risk minimization.
\newblock \emph{arXiv preprint arXiv:1710.09412}, 2017.

\bibitem[Zhang et~al.(2020)Zhang, Yu, Li, Song, Liu, and Dai]{WSOD}
Jing Zhang, Xin Yu, Aixuan Li, Peipei Song, Bowen Liu, and Yuchao Dai.
\newblock Weakly-supervised salient object detection via scribble annotations.
\newblock In \emph{2020 IEEE/CVF Conference on Computer Vision and Pattern Recognition (CVPR)}, 2020.

\bibitem[Zhang and Zhuang(2022)]{2022CycleMix}
Ke Zhang and Xiahai Zhuang.
\newblock Cyclemix: A holistic strategy for medical image segmentation from scribble supervision.
\newblock In \emph{Proceedings of the IEEE/CVF Conference on Computer Vision and Pattern Recognition (CVPR)}, pages 11656--11665, 2022.

\bibitem[Zhou et~al.(2022{\natexlab{a}})Zhou, Chen, Gui, Li, and Wang]{2022_split}
Gaodian Zhou, Weitao Chen, Qianshan Gui, Xianju Li, and Lizhe Wang.
\newblock Split depth-wise separable graph-convolution network for road extraction in complex environments from high-resolution remote-sensing images.
\newblock \emph{IEEE Transactions on Geoscience and Remote Sensing}, page 1–15, 2022{\natexlab{a}}.

\bibitem[Zhou et~al.(2018)Zhou, Zhang, and Ming]{2018D}
L. Zhou, C. Zhang, and W. Ming.
\newblock D-linknet: Linknet with pretrained encoder and dilated convolution for high resolution satellite imagery road extraction.
\newblock In \emph{2018 IEEE/CVF Conference on Computer Vision and Pattern Recognition Workshops (CVPRW)}, 2018.

\bibitem[Zhou et~al.(2022{\natexlab{b}})Zhou, Sui, Chen, Liu, Shi, and Chen]{Zhou_Sui_Chen_Liu_Shi_Chen}
Mingting Zhou, Haigang Sui, Shanxiong Chen, Junyi Liu, Weiyue Shi, and Xu Chen.
\newblock Large-scale road extraction from high-resolution remote sensing images based on a weakly-supervised structural and orientational consistency constraint network.
\newblock \emph{ISPRS Journal of Photogrammetry and Remote Sensing}, 193:\penalty0 234--251, 2022{\natexlab{b}}.

\bibitem[Zhu et~al.(2017)Zhu, Park, Isola, and Efros]{PatchGAN}
Jun-Yan Zhu, Taesung Park, Phillip Isola, and Alexei~A. Efros.
\newblock Unpaired image-to-image translation using cycle-consistent adversarial networks.
\newblock In \emph{2017 IEEE International Conference on Computer Vision (ICCV)}, 2017.

\end{thebibliography}
}
% \thanks{Manuscript received January 00, 0000; revised January 00, 0000; accepted January 00, 0000. This work was supported in part by the National Natural Science Foundation of China under Grant 00000000; in part by the Innovation Capability Support Program of Shaanxi (Program No. 2021KJXX-08), by the Xi'an Association for Science and Technology Youth Talent Support Program Project under Grant 095920201301, by the Fundamental Research Funds for the Central Universities under Grant JB211901, and by the Aeronautical Science Fund of China under Grant 2019ZC081002.}

\clearpage
\setcounter{page}{1}
\maketitlesupplementary

\section{Rationale}
\label{sec:rationale}
\subsection{Sensitivity Analysis of Regularization Weights}

\begin{table}[htbp]
  \centering
    \begin{tabular}{ll|cc}
    \toprule
         \multicolumn{2}{c|}{Loss Weight}   & \multicolumn{2}{c}{DeepGlobe} \\
          \midrule
    $\lambda_1(\mathcal{L}_{inv})$     & $\lambda_2(\mathcal{L}_{C-D})$    & IoU   & F1 \\
    \midrule
    0     & 0     & 59.19  & 74.37  \\
    0.01  & 0.01  & 59.70  & 74.76  \\
    0.05  & 0.05  & 59.61  & 74.69  \\
    \textbf{0.1}   & \textbf{0.1}   & \textbf{60.38}  & \textbf{75.29}  \\
    0.25  & 0.25  & 59.94  & 74.95  \\
    0.5   & 0.5   & 59.53  & 74.63  \\
    0.75  & 0.75  & 59.82  & 74.86  \\
    1     & 1     & 59.80  & 74.85  \\
    \bottomrule
    \end{tabular}%
    \caption{Sensitivity analysis of loss weights on DeepGlobe dataset}
  \label{tab:add_sens}%
\end{table}%

\cref{tab:add_sens} summarizes the obtained IoU and F1 scores by SA-MixNet on DeepGlobe dataset with different regularization weights. 
We denote the weights for $\mathcal{L}_{inv}$ and $\mathcal{L}_{C-D}$ as $\lambda_1$ and $\lambda_2$, respectively. 
When $\lambda_1$ and $\lambda_2$ are both set to 0.01, there is a noticeable improvement in IoU and F1 scores against that without $\mathcal{L}_{inv}$ and $\mathcal{L}_{C-D}$, \textit{i.e. $\lambda_1 = \lambda_2 = 0$}. 
This indicates that the designed invariance regularization and connectivity regularization can effectively enhance the model's performance. 
As the weight of regularization continues increasing, the model reaches its highest IoU and F1 scores at $\lambda_1 = 0.1$ and $\lambda_2 = 0.1$. 
While the weight of regularization is further increased, there may be a performance decline due to the potential trade-off effects on the base segmentation loss.

\subsection{Generalizability Experiment}
\begin{table}[h]
  \centering
    \begin{tabular}{c|cc|cc}
    \toprule
     Method & \multicolumn{2}{c|}{BPG} & \multicolumn{2}{c}{BPG$\ +\ $SA-MixNet} \\
    \midrule
    Dataset & IoU   & F1    & IoU   & F1 \\
    \midrule
    DeepGlobe & 54.00  & 70.13  & 59.01  & 74.22  \\
    Wuhan   & 50.86  & 67.43  & 54.19  & 70.29  \\
    Ma    & 55.45  & 71.34  & 60.62  & 75.49  \\
    \bottomrule
    \end{tabular}%
    \caption{Generalizability experiment based on BPG}
  \label{tab:BPG}%
\end{table}%

In \cref{subs:gen}, we discuss the framework's generalizability by applying SA-MixNet on ScRoadExtractor's DBNet \cite{2021Scribble} and present the performance on DeepGlobe, Wuhan, and Massachusetts datasets in \cref{tab:generalize}. 
To further validate the generalizability of the framework,  our SA-MixNet, as a learning framework, is applied to BGP \cite{BPG} and WeaklyOSM's MD-ResUnet(MD) \cite{2019RoadScribble}, and the proposed SC-label expansion, SA-Mixup and invariance-based regularization are integrated to them.

\cref{tab:BPG} presents the performance of applying SA-MixNet to BPG. 
The introduced new pseudo-labeling scheme and the learning framework lead to improvements of 5.01\%, 3.33\%, and 5.17\% in the IoU performance metric on DeepGlobe, Wuhan and Massachusetts datasets, respectively.

% Table generated by Excel2LaTeX from sheet 'trans_MD'
\begin{table}[htbp]
  \centering
    \begin{tabular}{c|cc|cc}
    \toprule
     Method & \multicolumn{2}{c|}{MD} & \multicolumn{2}{c}{MD$\ +\ $SA-MixNet} \\
    \midrule
    Dataset & IoU   & F1    & IoU   & F1 \\
    \midrule
    DeepGlobe & 54.32  & 70.40  & 58.44  & 73.77  \\
    Wuhan   & 52.34  & 68.72  & 54.46  & 70.52  \\
    Ma    & 52.74  & 69.06  & 61.97  & 76.54  \\
    \bottomrule
    \end{tabular}%
      \caption{Generalizability experiment based on MD-ResUnet}
  \label{tab:MD}%
\end{table}%

Similar performance improvements are also observed when applying SA-MixNet to MD-ResUnet.
As shown in \cref{tab:MD}, there are performance gains of 4.12\%, 2.12\%, and 9.23\% in the term of IoU scores on all selected datasets, respectively.
Overall, the application of the proposed framework (SA-MixNet) to DBNet, BPG, and MD-ResUnet leads to significant performance improvements, regardless of whether additional supervision or priors are used by the original methods or not.
This demonstrates the potential of our proposed framework to be generalized and applied to other methods, enhancing their performance by creating more diverse and complex samples, and forcing the models to behave invariantly on various scenes.

\end{document}